%% file: colm2025_conference.tex
\documentclass{article} 
\usepackage[preprint]{colm2025_conference}
\usepackage{microtype}
\usepackage{hyperref}
\usepackage{url}
\usepackage{booktabs}
\usepackage{amsmath}
\usepackage{lineno}
\usepackage[pdftex]{graphicx}

\definecolor{darkblue}{rgb}{0, 0, 0.5}
\hypersetup{colorlinks=true, citecolor=darkblue, linkcolor=darkblue, urlcolor=darkblue}

\title{Beyond Accuracy: The Role of Calibration in Self-Improving Large Language Models}

\author{
Liangjie Huang \\
University of Illinois Chicago \\
\texttt{lhuan85@uic.edu}
\And
Dawei Li \\
Arizona State University \\
\texttt{daweili5@asu.edu}
\And
Huan Liu \\
Arizona State University \\
\texttt{huanliu@asu.edu}
\And
Lu Cheng \\
University of Illinois Chicago \\
\texttt{lucheng@uic.edu}
}

\begin{document}


\maketitle

\begin{abstract}
Large Language Models (LLMs) have demonstrated remarkable self-improvement capabilities, whereby models iteratively revise their outputs through self-generated feedback. While this reflective mechanism has shown promise in enhancing task performance, recent studies suggest that it may also introduce undesirable biases—most notably, self-bias, or the tendency of LLMs to favor their own prior outputs. In this work, we extend this line of inquiry by investigating the impact on confidence estimation. We evaluate three representative self-improvement paradigms—basic prompting, Chain-of-Thought (CoT) prompting, and tuning-based methods—and find that iterative self-improvement can lead to systematic overconfidence, as evidenced by a steadily increasing Expected Calibration Error (ECE) and lower accuracy with high confidence.  We then further explore the integration of confidence calibration techniques with self-improvement. Specifically, we compare three strategies: (1) applying calibration after multiple rounds of self-improvement, (2) calibrating before self-improvement, and (3) applying calibration iteratively at each self-improvement step. Our results show that iterative calibration is most effective in reducing ECE, yielding improved calibration. Our work pioneers the study of self-improving LLMs from a calibration perspective, offering valuable insights into balancing model performance and reliability.

\end{abstract}

\section{Introduction}
\input{intro_v2}

\section{Related Work}


\textbf{Self-Improvement} generally refers to the way that LLMs try to review and correct their own mistakes to achieve performance improvement on their own.
Broader view on this topic can be categorized into three types of methods~\citep{kamoi2024can}: \textit{Intrinsic Improvement}, \textit{External Information} and \textit{Fine-tuning}. Intrinsic Improvement means LLMs generate feedbacks to their own responses and correct themselves~\citep{kim2023language,dhuliawala2024chain}.
Recently, some researchers found that intrinsic improvement can be affected by the prompting mechanism. Specifically, prompting with CoT and self-refinement style have gained effective results~\citep{shinn2023reflexion,madaan2023self,fu2025multiple}. 
An iterative and intrinsic self-improvement process where LLMs generates a response, receives feedback via a feedback model, and refines its output using the same model as a refinement model.
\textit{External information} will introduce some extra tools to help check the responses from LLMs. These include many scopes, such as code executors~\citep{chen2023teaching}, search engines~\citep{zhao2023verify}, human feedback~\citep{chen2024learning} and so on. \textit{Fine-tuning} for self-improvement generates feedback and then refines its responses, so that it achieves self-improvement via learning from these corpus. Popular methods in this branch are supervised fine-tuning~\citep{first2023baldur,han2024small,zhang2024small} and reinforcement
learning (RL)~\citep{akyurek2023rl4f,xie2025teaching}.


In this study, we focus on intrinsic self-improvement, a concept that has attracted considerable debate in recent years. On the one hand, studies such as \cite{bai2022constitutional} suggest that prompting LLMs can enable them to self-correct harmful outputs. Other work, including self-refine approaches \citep{madaan2023self} and RCI Prompting \citep{kim2023language}, demonstrates how LLMs can iteratively refine their own responses in reasoning tasks. On the other hand, \cite{huang2023large} indicates that LLMs may struggle to enhance their performance without external feedback, and that their performance can even degrade after self-improvement attempts. Further research similarly reports that achieving self-improvement by solely relying on prompts remains challenging \citep{gou2023critic, olausson2023self}. These findings motivate us to investigate the underlying mechanisms and conditions under which intrinsic self-improvement can be most effectively realized. Additionally, we also adopt supervised fine-tuning method for self-improvement, considering its efficiency and effectiveness compared with the RL-based one.


\textbf{Calibration.}
Popular methods for calibrating language models can be broadly classified into five categories: verbalization-based, self-consistency-based, logit-based, internal state-based, and surrogate approaches~\citep{geng2023survey,xie2024survey}. Verbalization-based methods leverage an LLM to explicitly express uncertainty about its answers. For instance, \cite{xu2024sayself} fine-tune its language models and then prompt the LLMs to indicate the confidence of its response by generating a probability scaler. Self-consistency-based methods rely on the intuition that confident models produce consistent outputs. Consequently, these methods sample multiple responses and estimate confidence by clustering outputs based on similarity~\citep{huang2024calibrating}. Internal state-based examines how the model’s internal layers (like attention heads or hidden states) respond during generation~\citep{azaria2023internal,li2023inference}. And surrogate models are used to mimic or approximate a black-box LLM in order to estimate confidence or uncertainty~\citep{shrivastava2023llamas}.

However, both verbalization-based and self-consistency-based methods may be constrained by the LM's ability to follow instructions accurately. 
Appropriate layers or heads in internal state-based methods vary a lot and thus are hard to unify for comparison.
And the surrogate model is not the same as the target model.
For a better fit to our research goal, logit-based methods directly utilize predicted token probabilities to evaluate response confidence~\citep{huang2023look}. Typically, logits are transformed or calculated to represent the forecasted confidence. The logits are believed to have the capacity to offer a more nuanced understanding of confidence knowledge  \citep{widmann2021calibration,kuhnsemantic,jang2024calibrated}. Notably, as a logits-based method, temperature scaling has been widely applied in LLMs for answering questions. By adjusting the temperature parameter, it influences the model's probability distribution over possible answers, thereby enhancing its performance in selecting the correct option~\citep{peeperkorn2024temperature,xie2024calibrating,shen2024thermometer}. In this work, we thus use logits-based calibration approach to discover the relationship between self-improvement and calibration in multi domains.

\section{Methods}
In this section, we introduce the three backbone techniques for self-improvement, as well as various manners to marry self-improvement methods with calibration. The overall framework can be found in Figure \ref{fig:overview}.

\subsection{Self-Improvement}



\textbf{Basic Prompting.} Basic prompting in this work refers to clearly and directly prompt LLMs to answer questions without guiding the LLM to output its CoT. As shown in left panel of Figure ~\ref{fig:overview}, an LLM generates an answer $a^{(0)} =\mathrm{LLM}(\mathrm{~}q\mathrm{~})$ given query $\mathrm{~}q\mathrm{~}$. Subsequently, this initial answer undergoes an evaluation phase, where feedback $f^{(t)}$ in round $t$ is produced. This feedback is then in a subsequent step "Answer Refining," to revise the answer given the feedback to get $a^{(t+1)}$.
\begin{equation}\begin{aligned}
f^{(t)} & =\mathrm{LLM}(q,\mathrm{~}a^{(t)}),\quad t\geq0;
\quad
a^{(t+1)} & =\mathrm{LLM}(q,\mathrm{~}a^{(t)},\mathrm{~}f^{(t)}).
\end{aligned}\end{equation}

This iterative loop above involving answer generation, feedback provision, and refinement contributes to enhancing the LLM's performance~\citep{madaan2023self}.

\textbf{Chain-of-Thought (CoT) Prompting.} CoT Prompting~\citep{wei2022chain} involves guiding LLMs through step-by-step reasoning to solve complex problems, generate detailed explanation or feedback. Recent advancement, including the emerging DeepSeek-R1 models~\citep{guo2025deepseek}, leverage CoT by explicitly generating structured intermediate steps, significantly boosting the reasoning capabilities. We also introduce this technique as one of our self-improvement methods. Instead of directly outputting answers from LLMs, we first guide the language models to generate a CoT response $c=\mathrm{LLM}_{\mathrm{CoT}}(\mathrm{~}q\mathrm{~})$, explicitly articulating the reasoning steps involved in answering the query. After getting the CoT for a specific question, we then use this generated CoT as new context to guide the LLM to provide the answer $a =\mathrm{LLM}(q,\mathrm{~}c)$: 
\begin{equation}\begin{aligned}
f^{(t)} & =\mathrm{LLM}(q,\mathrm{~}c,\mathrm{~}a^{(t)}),\quad a^{(t+1)}=\mathrm{LLM}(q,\mathrm{~}c,\mathrm{~}a^{(t)},\mathrm{~}f^{(t)}).
\end{aligned}\end{equation}


\textbf{Supervised Fine-Tuning (SFT).} Apart from prompting, we also utilize SFT~\citep{dong2024abilities,dong2024threshold} with specific datasets to investigate the resoning ability in LLMs, thereby exploring its role in self-improvement and calibration. A SFT loss is typically defined as

\begin{equation}\mathcal{L}_{\mathrm{SFT}}(\theta)=-\sum_{(q,y)\in\mathcal{D}}\sum_{i=1}^I\log p_\theta(y_i|q,y_{<i}).\end{equation}

Noted that $I$ is the total number of tokens in the target output sequence $y$. And $p_\theta(y_i|q,y_{<i})$ is the probability assigned by the model to token 
$y_i$. Query $q$ is input sequence, which means the question prompt. $i$ is to locate output sequence position. $\theta$ is the model parameters. $\mathcal{D}$ is the finetuning dataset, a collection of question query and the according answer pairs.


\subsection{Calibration}
In the context of LLMs, calibration refers to how well an LLM's predicted confidence aligns with its actual accuracy.
As one of the most common calibration approaches, temperature scaling~\citep{guo2017calibration} is a post-hoc calibration strategy that aligns model predictions with observed probabilities. We adapt the method from~\citep{shen2024thermometer}, a temperature scaling calibration approach tailored to LLMs that learns an auxiliary model to map the outputs of the LLM to better-calibrated probabilities. The calibration formula is shown below: 
\begin{equation}p(y_n|q_n,\tau_k;W)=\frac{\exp(w_y^T\phi(q_n;W)/\tau_k)}{\sum_{v^{\prime}}\exp(w_{v^{\prime}}^T\phi(q_n;W)/\tau_k)}.\end{equation}
The key idea is to train an neuro network to fit the logits distribution and then use the network to infer task-specific latent temperatures $\tau$, allowing the model to adapt to new questions with learned parameters. $\phi(q_n;W)$ is the feature that the language model produces for the input token sequence $q_n$. $\sum_{v^{\prime}}\exp(w_{v^{\prime}}^T\phi(q_n;W)/\tau_k)$ is the sum of exponential over all possible tokens $v^{\prime}$ in the vocabulary. $W$ and $w$ are model parameters and the logit vector transformation, respectively. The method is computationally efficient, to preserve the accuracy of the LLM, and takes a step towards being universal among different tasks.

\subsection{Marrying Self-improvement with Calibration}


We propose three methods to answer the second research question, via marrying self-improvement with calibration, as shown in Figure ~\ref{fig:overview}. 

\textbf{The Iterative Method} refers to a process where each round consists of basic-prompting-based self-improvement followed by calibration. It facilitates a direct observation of how self-improvement and calibration mutually enhance or constrain each other during successive iterations.

\begin{equation}a^{(t+1)}=\mathrm{~Calibrate}{\left(\text{Self-Improve}(a^{(t)})\right)},\quad t=0,1,2,\ldots\end{equation}

In the equation, $a^{(0)}$ is the inital response from LLM for query $q$ and $a^{(t+1)}$ is the result after self-improvement and calibration in each round.

\textbf{Calibration then Self-improvement} performs calibration only once at the beginning, and in the subsequent rounds, only self-improvement is conducted. 
\begin{equation}a^{(0)}=\mathrm{Calibrate}{\left(a^{(0)}\right)}; \quad
a^{(t+1)}=\text{Self-Improve}{\left(a^{(t)}\right)},\quad t=0,1,2,\ldots\end{equation}
This helps determine if an initial calibration provides a stronger foundation for subsequent self-improvements, reducing the risk of deviation from the ideal state.

In \textbf{Multi Self-Improvement then Calibration}, it instructs the LLM to perform $T$ rounds of self-improvement first, followed by a single calibration: 
\begin{equation}a^{(t+1)}=\text{Self-Improve}{\left(a^{(t)}\right)},\quad t=0,1,2,\ldots,T-1. \quad a^{\mathrm{(final)}}=\mathrm{Calibrate}{\left(a^{(T)}\right)}.\end{equation}
This design allows the model to freely explore and maximize its potential before using calibration to correct accumulated errors and biases. It also helps to assess whether a single calibration step remains effective in correcting deviations accumulated through multiple self-improvement iterations.


\section{EXPERIMENTS}

\subsection{Set Up}
\textbf{Models.} In this paper, we use popular open-source LLMs. Specifically, Llama-2-7b-chat-hf~\citep{touvron2023llama} as a standard LLM and DeepSeek-R1-Distill-Llama-8B~\citep{guo2025deepseek} as a deep thinking model, denoted as Llama-deepseek in later section, are used to investigate the effectiveness and relationship between self-improvement and calibration. 

\textbf{Dataset.} MMLU~\citep{hendrycks2020measuring} is utilized in our paper for evaluation. The MMLU is a comprehensive benchmark, which covers 57 sub-datasets spanning various subjects including STEM (Science, Technology, Engineering, and Mathematics), humanities, social sciences, and other specialized areas. It consists of multi-domain questions that assess both world knowledge and problem-solving abilities, making it well-suited for evaluating calibration and self-improvement. We use all 57 sub-datasets for the experiments.

In terms of SFT dataset, we follow \citep{zhang2024small}, which focuses on the self-correction abilities of small, open-source LMs, exploring whether they can self-correct with minimal guidance. 
We adopt their refined dataset to fine-tune the LLMs.

\textbf{Evaluation Metrics.}
We investigate our research using these two metrics: Accuracy (ACC) for LLM prediction accuracy and ECE for model calibration measurement. ECE is a widely adopted metric that measures the discrepancy between predicted confidence and its actual accuracy. A high ECE score reflects poor calibration, indicating a significant discrepancy between the model's predicted confidence and its empirical accuracy on the given dataset.

\begin{equation}\mathrm{ECE}=\sum_{k=1}^{K}\frac{|B_{k}|}{N}\left|\mathrm{acc}(B_{k})-\mathrm{conf}(B_{k})\right|.\end{equation}

ECE essentially computes a weighted sum of absolute differences between accuracy \(\text{acc}(B_k)\) and confidence \(\text{conf}(B_k)\) across bins. Here, we use 10 bins of width 0.1 each in [0,1], where N denotes the number of model's generations and $K$ denotes the number of bins.

\textbf{Supervised Fine-Tuning (SFT).}
We performed SFT of the base models using Low-Rank Adaptation (LoRA)~\citep{hu2022lora}. Specifically, the LoRA configuration employed a rank ($r$) of 32, along with a scaling factor (lora\_alpha) of 16. LoRA dropout was set to 0.05. Training was conducted using a batch size of 8. To efficiently manage GPU memory, gradient checkpointing was enabled. The maximum gradient norm was clipped at 0.3 to ensure training stability. We fine-tuned the model for five epochs using a learning rate of 2e-4, coupled with a cosine learning rate scheduler and a warm-up ratio of 0.05. Additionally, training with BF16 was utilized to enhance training efficiency. One A100 GPU was used in our experiment.

\subsection{Results and Analysis}
In this section, we will be answering the following research questions introduced in the Introduction with the experiment results analysis.

\begin{itemize}
    \item \textbf{RQ 1:} Will self-improvement lead to LLM self-bias in confidence estimation?
    \item \textbf{RQ 2:} What are the compounded effects of marrying calibration and self-improvement on model performance?
\end{itemize}
\subsubsection{RQ 1: Will self-improvement lead to bias in confidence estimation?}



\begin{figure}[htbp]
    \centering
    \includegraphics[width=\textwidth]{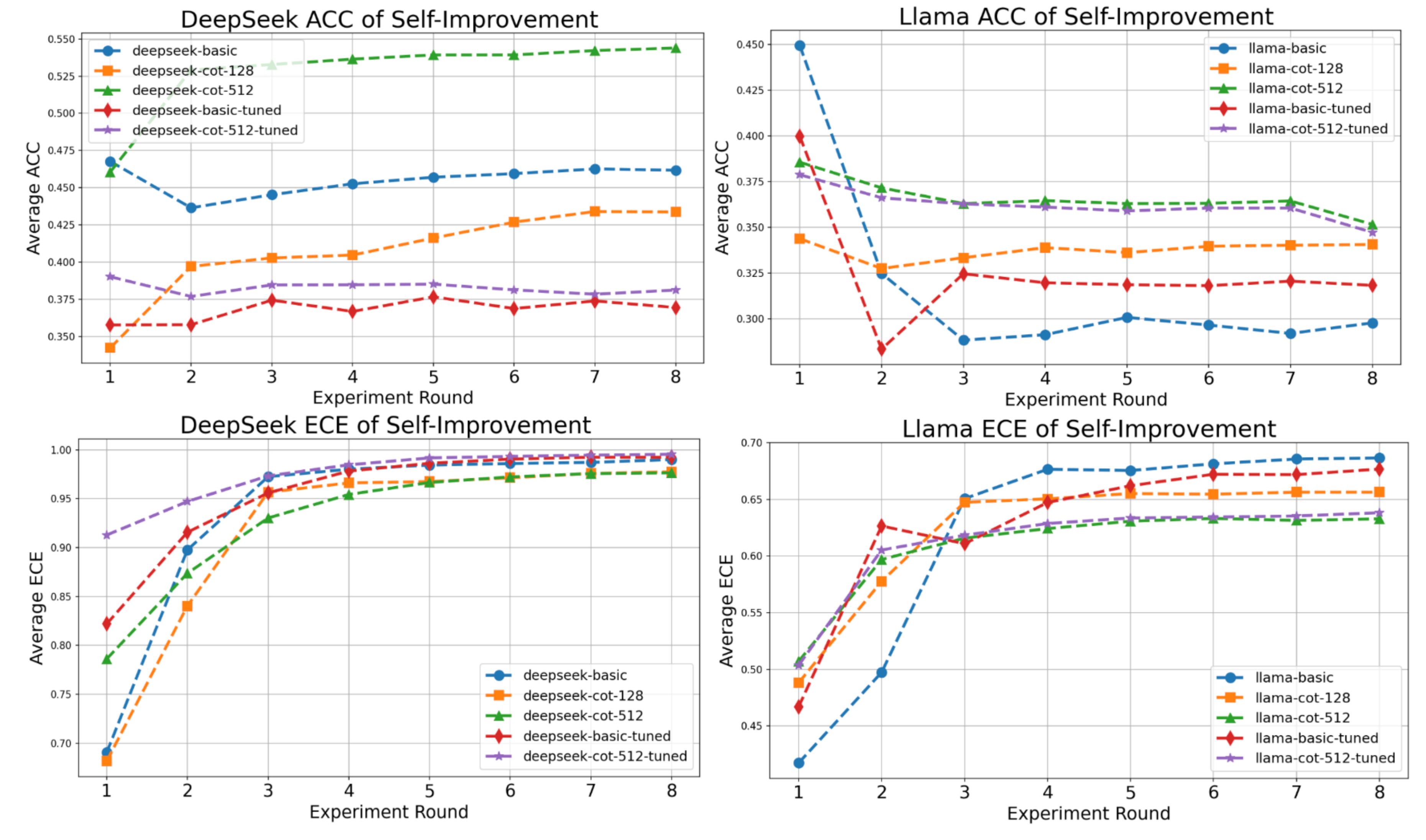}
    \caption{Results of Self-Improvement in Different Methods.}
    \label{fig:si}
    {\footnotesize \textit{Note.} Basic means the basic prompting method and cot is for CoT prompting with different length of tokens. Tuned stands for the fine-tuned method}
\end{figure}

As seen in Figure ~\ref{fig:si}, the two upper charts are the accuracy scores of Llama-deepseek and Llama, on the left and right respectively. Similarly, the two bottom charts are the ECE scores. 



\textbf{Longer CoT generally enhances model accuracy but model’s inherent reasoning capacity can modulate its effectiveness.} Longer CoT consistently yielded the highest accuracy across both Llama-deepseek and Llama, with Llama-deepseek demonstrating a clear trend of progressive self-improvement over multiple rounds~\citep{jin2024impact}. While Llama’s accuracy eventually declined after several rounds of self-improvement using a longer CoT, it still outperformed other self-improvement methods within Llama. Notably, the CoT methods with 512 tokens in Llama experienced a late drop in accuracy due to the 4096-token context limitation. Moreover, CoT length significantly influenced inference accuracy: longer CoT (512-token limit) reliably produced higher accuracy compared to shorter CoT (128-token limit). Interestingly, for Llama—the weaker of the two base models—shorter CoT sequences provide a moderate boost, suggesting limited CoT can still benefit models of relatively constrained reasoning capabilities.

\textbf{The effectiveness of prompting-based methods appears to be strongly influenced by the model’s intrinsic reasoning capabilities.}  In Llama-deepseek, while basic prompting experienced a slight decline in accuracy initially, it subsequently facilitated continuous error correction and progressive improvement. By contrast, Llama exhibited a general deterioration in ACC as the number of prompting rounds increased, with basic prompting ultimately yielding the lowest accuracy among all tested self-improvement strategies.

\textbf{SFT may be more beneficial for weaker models.} Our fine-tuning experiments revealed divergent outcomes in these two models. In Llama-Deepseek, both CoT with 128 and 512 tokens exhibited lower ACC than the original basic prompting method and the fine-tuning ones. Notably, fine-tuned basic prompting resulted in the poorest ACC among all conditions, with one of the highest ECE. This indicates that calibration worsened for Llama-deepseek post-fine-tuning. In Llama, however, fine-tuning produced improvements: both fine-tuned methods surpassed the original basic prompting in terms of ACC, and their ECE also improved, suggesting better calibration. The performance drop in Llama-deepseek after fine-tuning may stem from a mismatch between the fine-tuning dataset and the reasoning dataset, thereby causing noticeable degradation. As Llama-deepseek possesses stronger inherent reasoning capabilities, it may be ill-suited to the chosen dataset and approach. In contrast, Llama, with weaker intrinsic reasoning ability, appears to benefit from fine-tuning, which leads to more pronounced gains.

In addition, Llama-deepseek exhibited a substantially higher initial ECE than Llama and maintained an high level in all self-improvement experiments, suggesting poor calibration under iterative improvement. In contrast, although basic prompting yielded the highest ECE among Llama’s self-improvement methods, its ECE does not exceed 0.7--lower than that of Llama-deepseek--indicating that Llama remains inherently better calibrated than Llama-deepseek. Moreover, ECE values tended to be lower when CoT reasoning was applied, particularly in Llama. To further investigate these findings, we propose conducting confidence distribution bias experiments to compare predicted confidence versus actual accuracy across various confidence intervals (e.g., 0.1–0.2, 0.2–0.3, etc.) under both basic and CoT prompting for the two LLMs.

\paragraph{Self-Bias in Confidence Estimation.}

\begin{figure}[htbp]
    \centering
    \includegraphics[width=1.0\textwidth]{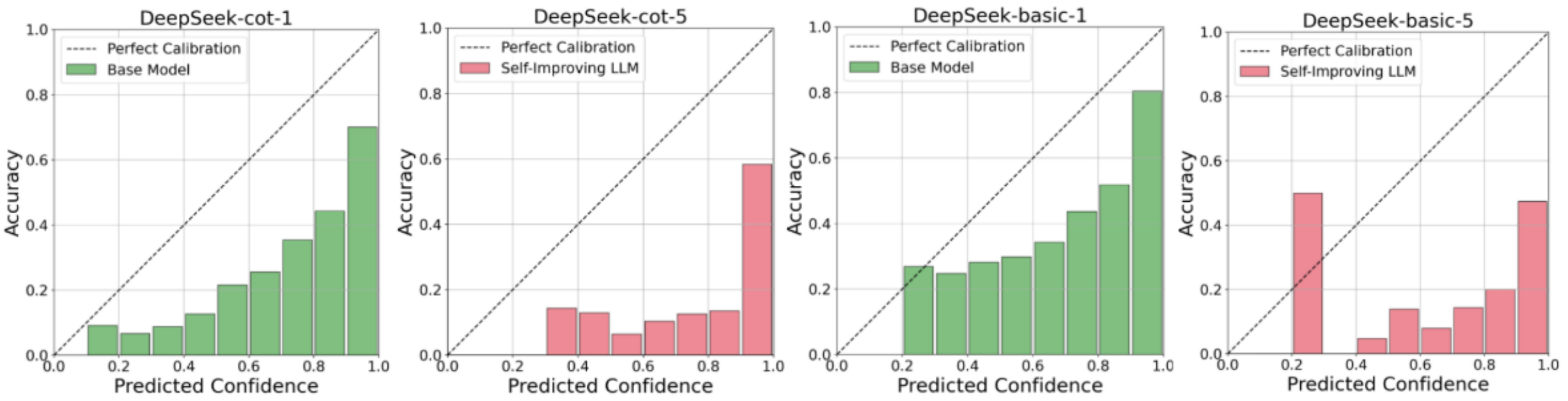}
    \caption{Llama-deepSeek's accuracy and confidence distribution.}
    \label{fig:ds_acc_conf}
\end{figure}

We use the longer CoT and basic prompting to illustrate self-improvement performance at the initial and intermediate stages. The x-axis in Figure~\ref{fig:ds_acc_conf} represents confidence levels divided into ten bins, while the y-axis denotes the corresponding accuracy.

We can observe that in Llama-deepseek, there is no substantial calibration improvement from the first to the fifth round; rather, the performance appears to deteriorate. In particular, the accuracy of high-confidence predictions decreases, suggesting that self-improvement might have exacerbated overconfidence in certain areas. Furthermore, during Llama-deepseek’s self-improvement with basic prompting, a notable dip in confidence occurs in the 0.2–0.3 interval. 



\begin{figure}[htbp]
    \centering
    \includegraphics[width=1.0\textwidth]{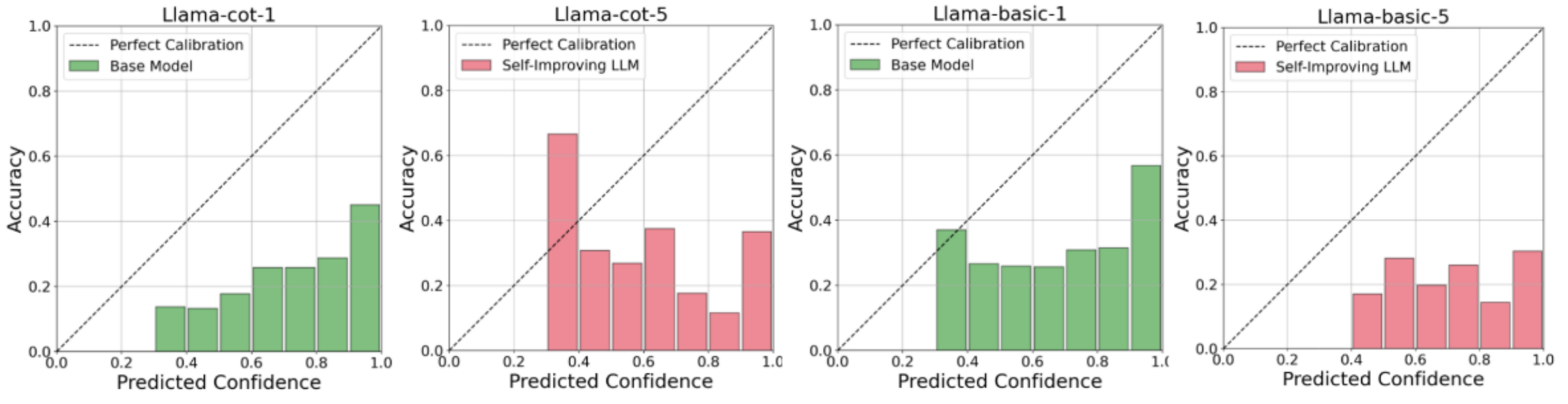}
    \caption{Llama's accuracy and confidence distribution.}
    \label{fig:llama_acc_conf}
\end{figure}

In Figure~\ref{fig:llama_acc_conf}, we observe that in the initial round for Llama, the relationship between confidence and accuracy generally aligns with expectations: as confidence increases, accuracy also improves. However, in the high-confidence region, the model exhibits a tendency toward overconfidence, characterized by high confidence yet relatively lower accuracy. By the fifth round of self-improvement, this issue becomes more pronounced, exacerbating the overconfidence effect. Similarly, during the fifth round of CoT, we observe a sudden rise in accuracy within the 0.3–0.4 confidence range. 
Based on these, we thus can conclude that \textbf{prompting and fine-tuning based methods in iterative self-improvement can introduce or amplify self-biases in confidence estimation}.



\subsubsection{RQ 2: What are the compounded effects of marrying calibration and self-improvement on model performance?}
\begin{figure}[htbp]
    \centering
    \includegraphics[width=1.0\textwidth]{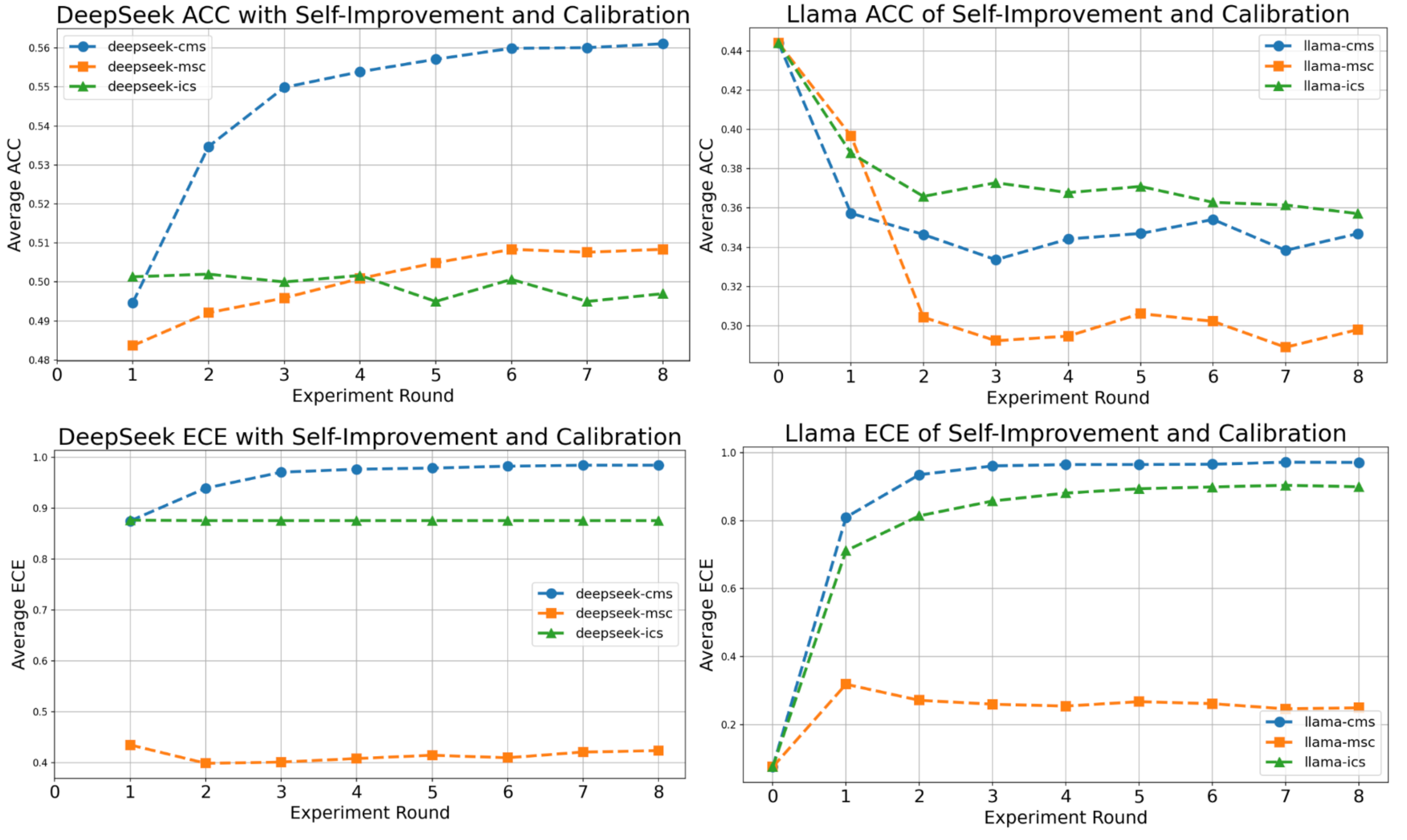}
    \caption{Self-Improve and Calibration Relationship Experiment Result.}
    \label{fig:cali}
    {\footnotesize \textit{Note.} Cms means calibration then multi self-improvement and msc is multi self-improvement then calibration. Ics stands for iterative calibration and self-improvement}
\end{figure}


As calibration serves as an effective technique to align a model’s confidence with its correctness and thus improve models confidence estimation, we propose three experiments using the basic prompting approach to investigate the RQ2. The results highlight notable commonalities and distinctions between the LLMs, as shown in Figure ~\ref{fig:cali}.

\textbf{ECE can be diminished when combined with self-improvement after calibration.}
Multi self-improvement–then–calibration methods yield reduced ECE, with the latter achieving a markedly lower ECE compared to the other two approaches. Despite performing calibration after each round, the iterative method continues to exhibit relatively high ECE, possibly because the alternating introduction of self-improvement dilutes the calibration effect and consequently compromises alignment between confidence and accuracy. Furthermore, both the “calibration then multi self-improvement” and “iterative” methods produce relatively high ECE—particularly in Llama, where ECE increases substantially compared to self-improvement alone. One explanation for this phenomenon is that calibration is primarily intended to align the model’s confidence with its actual accuracy. However, during self-improvement, the model refines its responses based on self-generated feedback, which can shift its confidence distribution. As a result, the self-bias dominates over calibration effect when the calibration is performed at the beginning.

\textbf{Calibration can serve as a better foundation in self-improvement for stronger LLM.} The “calibration then multi self-improvement” strategy in Llama-deepseek shows steady improvement of ACC, surpassing the performance of Llama-deepseek’s longer CoT in pure self-improvement setting. Additionally, unlike basic prompting–based self-improvement, this method does not exhibit an initial accuracy drop. In Llama-deepseek, the multi self-improvement–then–calibration approach effectively rectifies errors in earlier stages while maintaining a relatively stable ECE; however, it manifests some fluctuations of ACC in later stages, suggesting a pronounced impact on model reasoning ability. Meanwhile, for Llama, the iterative method achieves the highest ACC across multiple rounds, although the overall trend still declines, reinforcing the notion that calibration is beneficial for self-improvement but Llama’s comparatively weaker intrinsic reasoning limits its capacity for effective self-correction.

\section{Conclusion}

In this work, we study the effect of self-improving LLM from a calibration perspective.
The first research question we propose is will self-improvement leads to self-bias in confidence estimation.
Based on our experiment results on three mainstream self-improvement approaches, we reveal an obvious trend of increasing overconfidence as self-improvement iterations progress, leading to a large ECE score value after the self-improvement process.
This motivates our second research question on how to marry calibration with self-improvement to mitigate this overconfidence.
With several potential solutions proposed and analyzed, we conclude that ECE can be largely diminished when applying calibration after self-improvement.
In the future, we will explore the calibration of self-improving LLMs in larger sizes of LLMs, as well as use a wider spectrum of calibration methods to validate the robustness and generalization of our findings. 
Besides, investigating self‐improvement and calibration in multilingual or multimodal settings would provide a richer understanding of how overconfidence manifests in more complex scenarios.

\bibliography{colm2025_conference}
\bibliographystyle{colm2025_conference}


\end{document}

%% file: intro_v2.tex
The development of Large Language Models (LLMs) has catalyzed transformative changes across numerous domains, from natural language understanding and generation~\citep{storks2019commonsense,weld2022survey} to assisting in complex question-answering and decision-making processes~\citep{li2025system,li2024generation,tan2024large}. To handle this, one of the emerging techniques for LLMs is self-improvement \citep{bai2022constitutional,kim2023language}, wherein LLMs iteratively review their own responses and refine their outputs based on self-generated feedback to enhancing the performance.
This process fosters human-like reflective thinking and has proven effective across a range of tasks and applications~\citep{tong2024can,pan2024automatically,li2024smoa}.

However, some recent studies also report cases where LLM-based self-improvement does not bring a significant boost and can even degrade the model's performance~\citep{zhang2024understanding,wu2024progress}.
One contributing factor to this counterintuitive outcome is self-bias~\citep{xu2024pride,wataoka2024self,li2025preference}—the tendency of LLMs to favor their own generated content. This cognitive bias impedes LLMs from providing impartial feedback on their outputs, thereby hindering effective self-correction and self-improvement.

\begin{figure}[htbp]
    \centering
    \includegraphics[width=.9\textwidth]{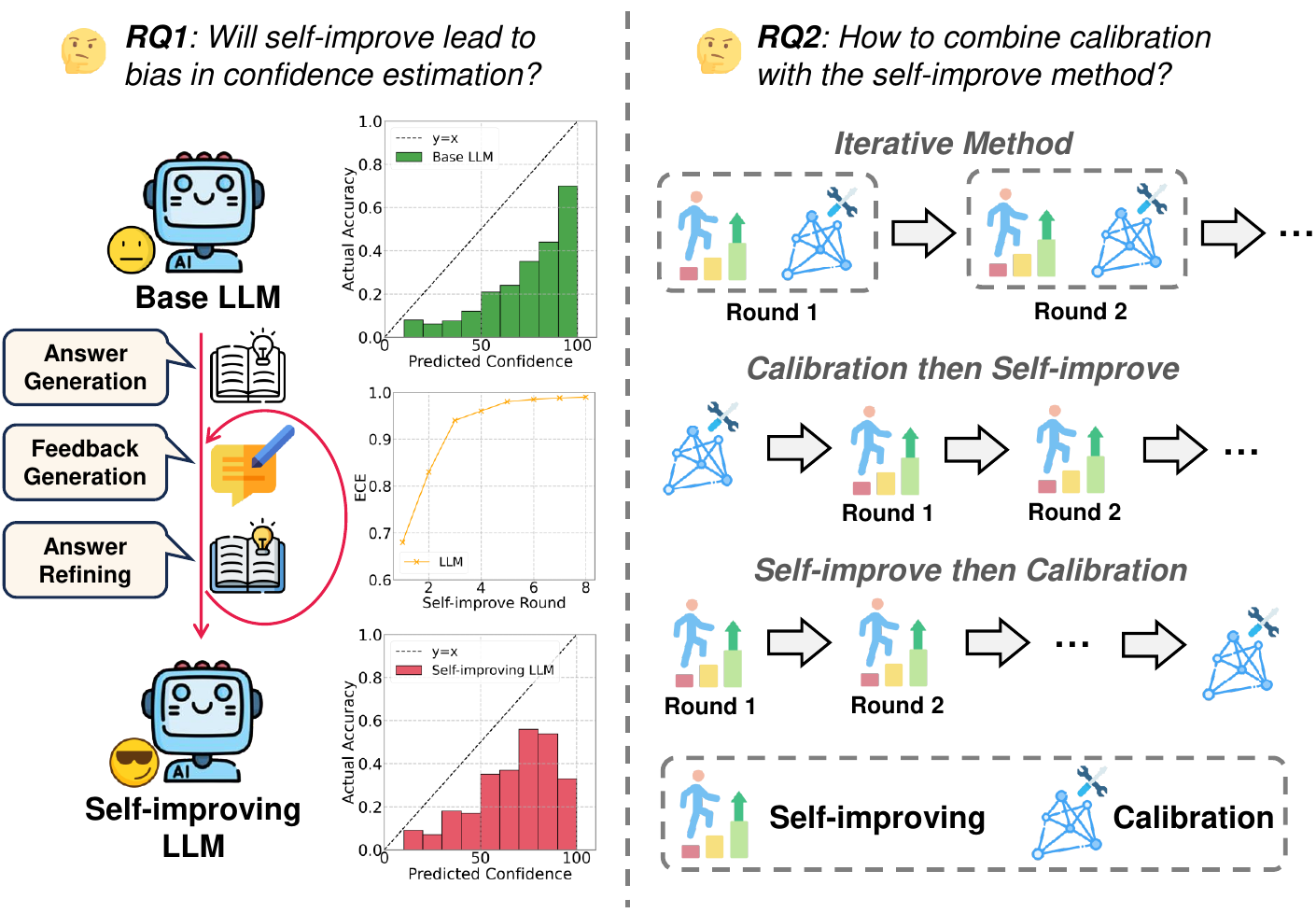}
    \caption{The two research questions and overview of our exploration process in this work.}
    \label{fig:overview}
\end{figure}

Borrowing this insight, we propose our \textbf{first} research question: \textit{Will self-improvement also lead to bias in confidence estimation?}
As LLMs become increasingly integral to both research and industry applications~\citep{zhu2025deepreview}, the ability to accurately express confidence or uncertainty in their outputs is crucial \citep{su2024api}, particularly in high-risk scenarios~\citep{thirunavukarasu2023large,li2024dalk}. If self-improvement methods introduce self-bias in confidence estimation, this could pose a significant threat to LLM safety and reliability, creating substantial challenges in the pursuit of trustworthy AI~\citep{sun2024trustllm,huang2025trustworthiness}.
To investigate this, we examine three types of self-improvement methods in our experiments: \textbf{Basic prompting}, \textbf{Chain-of-Thought (CoT) prompting}, and \textbf{Tuning-based approaches}~\citep{first2023baldur,han2024small,zhang2024small,akyurek2023rl4f,xie2025teaching}. We implement each method and analyze its impact on LLMs’ confidence estimation performance. Our results reveal a clear trend of increasing overconfidence as self-improvement iterations progress, leading to a continuously rising Expected Calibration Error (ECE) score \citep{guo2017calibration}.

As calibration~\cite{guo2017calibration,geng2023survey,xie2024survey} serves as an effective technique to align a model’s confidence with its correctness and thus improve models confidence estimation, we pose our \textbf{second} research question: \textit{How to combine calibration with the self-improvement method to mitigate the confidence estimation bias?}
To explore this, we examine the compounded effects of calibration and self-improvement. Specifically, we evaluate three experimental settings to analyze their interaction: (1) \textbf{multiple self-improvement iterations followed by calibration}, (2) \textbf{calibration applied before multiple self-improvement iterations}, and (3) \textbf{iterative calibration and self-improvement at each step}. Our results indicate that applying calibration before self-improvement leads to sustained improvements over time. Meanwhile, self-improvement then calibration achieves the best ECE score, resulting in better-calibrated confidence estimates. 

To summarize, our contribution in this paper is in two-fold:
\begin{itemize}
    \item From a novel perspective of calibration, we first propose to explore self-improvement's impact on LLMs' confidence estimation and reveal a significant overconfidence issue caused by iterative self-improvement.
    \item We explore several self-improvement paradigms to showcase the compounded effect when combining self-improvement with calibration, producing LLMs that are both effective and reliable in real-world applications.
\end{itemize}